%% file: acl2023.tex
\pdfoutput=1

\documentclass[11pt]{article}

\usepackage[final]{ACL2023}

\usepackage{times}
\usepackage{latexsym}
\usepackage{amsmath}
\usepackage{amssymb}
\usepackage{booktabs}
\usepackage{multirow}
\usepackage{graphicx}
\usepackage{colortbl}
\usepackage{subfig}
\usepackage{makecell}
\usepackage{bm}
\usepackage{bbm}
\usepackage{pgfplots}
\usepackage{lipsum}
\usepackage{float}
\pgfplotsset{compat=1.18}

\usepackage[T1]{fontenc}

\usepackage[utf8]{inputenc}

\usepackage{microtype}
\usepackage{xspace}

\usepackage{inconsolata}
\usepackage[ruled,linesnumbered]{algorithm2e}
\usepackage{algorithmic}
\definecolor{ForestGreen}{RGB}{34,139,34}
\definecolor{Gray}{gray}{0.9}
\definecolor{MyLightThemeColor}{RGB}{223, 213, 230}

\definecolor{MyDarkThemeColor}{RGB}{155, 128, 171}
\definecolor{MyDarkBlue}{RGB}{100, 126, 151}
\definecolor{MyLineChartColor}{RGB}{155, 128, 171}
\newcommand\DBKD{Ours\xspace}

\title{Bridging the Gap between Decision and Logits  in Decision-based \\ Knowledge Distillation for Pre-trained Language Models}

\author{
    Qinhong Zhou$^{1,3}$, Zonghan Yang$^{1,3}$, Peng Li$^{2,4,\dagger}$, Yang Liu$^{1,2,3,4,\dagger}$ \\
    $^1$Dept. of Comp. Sci. \& Tech., Institute for AI, Tsinghua University, Beijing, China \\
    $^2$Institute for AI Industry Research (AIR), Tsinghua University, Beijing, China \\
    $^3$Beijing National Research Center for Information Science and Technology \\
    $^4$Shanghai Artificial Intelligence Laboratory, Shanghai, China \\
}
\begin{document}
\maketitle
{\let\thefootnote\relax\footnotetext{$\dagger$ Peng Li (lipeng@air.tsinghua.edu.cn) and Yang Liu (liuyang2011@tsinghua.edu.cn) are corresponding authors.}}
\begin{abstract}
\input{sections/abstract.tex}
\end{abstract}
\input{sections/intro.tex}

\input{sections/related.tex}

\input{sections/method.tex}

\input{sections/exp.tex}

\input{sections/conclusion.tex}

\input{sections/limitations.tex}

\input{sections/ethics.tex}

\input{sections/acknowledgement}

\bibliography{anthology,custom}
\bibliographystyle{acl_natbib}

\input{sections/appendix.tex}

\end{document}

%% file: sections/abstract.tex
Conventional knowledge distillation (KD) methods require access to the internal information of teachers, e.g., logits. However, such information may not always be accessible for large pre-trained language models (PLMs). In this work, we focus on decision-based KD for PLMs, where only teacher decisions (i.e., top-1 labels) are accessible.
Considering the information gap between logits and decisions, we propose a novel method
to estimate logits from the decision distributions.
Specifically, decision distributions can be both derived as a function of logits theoretically and estimated with test-time data augmentation empirically. By combining the theoretical and empirical estimations of the decision distributions together, the estimation of logits can be successfully reduced to a simple root-finding problem.
Extensive experiments show that our method significantly outperforms strong baselines on both natural language understanding and machine reading comprehension datasets.\footnote{Our code is available at \href{https://github.com/THUNLP-MT/DBKD-PLM}{https://github.com/THUNLP-MT/\\ DBKD-PLM}}

%% file: sections/intro.tex
\section{Introduction}
Various natural language processing (NLP) tasks have witnessed promising performance from large pre-trained language models (PLMs)~\cite{bert, roberta, t5, gpt3}.
However, PLMs are usually computationally expensive and memory intensive, hindering their deployment on resource-limited devices.
Knowledge distillation (KD)~\cite{hintonkd} is a popular technique to transfer knowledge from large PLMs to lightweight models.
Previous KD works utilize various types of internal information from the teacher model, such as output logits~\cite{distilbert, tang2019distilling, fastbert}, hidden states~\cite{patientkd, tinybert}, and attention maps~\cite{emdkd}. In real-world applications, however, these types of information are sometimes not accessible due to commercial and privacy issues~\cite{gpt3,instructgpt}.
Specifically, large-scale PLMs usually
only provide \textit{decisions} (i.e., top-1 labels) to users.
Motivated by this scenario, we investigate the task of \textit{decision-based} KD~\cite{wang2021dbkd} for PLMs, in which only decisions of teacher predictions are available.



The information gap between teacher decisions and its internal states is the major challenge for the task. 
A straightforward approach for decision-based KD is to treat teacher decisions as ground truth labels and use these labels to train a student model~\cite{qekd,hardstealing}. 
However, previous work reveals that logits contain rich knowledge~\cite{hintonkd}, relying only on decisions obviously suffers from information loss. 
To alleviate the problem, \citet{wang2021dbkd} proposes the DB3KD method to generate pseudo soft labels according to the sample’s robustness. However, DB3KD requires that the input of a model can be modified continuously (e.g., image), which hinders its application on PLMs as their inputs are discrete tokens.
Therefore, how to fill the information gap under the discrete input setting remains a challenging problem.

Fortunately, the development of test-time data augmentation for discrete input~\cite{liu2019anonymizedbert, shleifer2019ulmfit, xu2022situ} brings hope for resolving the challenge. The basic idea is to modify selected tokens in a piece of text under certain constraints to generate augmented samples and estimate or improve the desired properties of a model based on its behaviors on these samples. Test-time data augmentation has been shown to be effective for uncertainty estimation~\cite{ayhan2018tta, smith2018tta, test-time-aug}, adversary robustness
~\cite{xu2022situ}, and so on.
Is it possible to narrow down the information gap with test-time data argumentation in decision-based KD for PLMs?


In this work, we propose a novel decision-based KD method for PLMs.
As illustrated in Figure~\ref{fig:framework}, our method is capable of estimating the teacher logits for classes even without observed decisions, narrowing down the information gap between decision and logits.
Specially, we estimate the logits by combining test-time data argumentation and non-centred orthant probability estimation.
On the one hand, we can obtain an empirical estimation of the decision distribution around a sample by test-time data argumentation. On the other hand, we can also derive a theoretical formula for the decision distribution as a non-centred orthant probability, which is a function of logits. As a result, the problem of logits estimation can be reduced to finding the root of the equation that the function takes the value of the empirical estimation.
Extensive experiments on various natural language understanding and machine reading comprehension datasets demonstrate the effectiveness of our proposed method, which outperforms strong baselines significantly. Moreover, quantitative analysis reveals that our method obtains better estimation of logits, narrowing down the information gap.




%% file: sections/related.tex
\section{Related Work}

\paragraph{Decision-based Knowledge Distillation.}
To advance conventional knowledge distillation (KD) to more challenging black-box model scenarios, \citet{wang2021dbkd} first propose the problem of decision-based KD, where only teacher decisions (i.e., top-1 labels) are accessible to students. They address the problem by estimating the soft label (analogy to output probabilities) of a sample based on its distance to the decision boundary, which involves continuous modification to the original input fed to the black-box model. Instead, \citet{qekd} and \citet{hardstealing} synthesize pseudo data in continuous space and leverage decisions of the teacher model on these data directly. In this work, we focus on decision-based KD for PLMs. Unfortunately, the original inputs of the PLMs are discrete tokens which can not be continuously modified. Therefore, these methods are not applicable to our scenario.

Decision-based KD is also related to black-box KD~\cite{Orekondy_2019_CVPR,Wang_2020_CVPR} and distillation-based black-box attacks~\cite{Zhou_2020_CVPR,Wang_2021_CVPR,Truong_2021_CVPR,Kariyappa_2021_CVPR,Yu-2022-FE-DaST}. Both of them involve distilling a student model from black-box models. However, these works generally assume that the score-based outputs are accessible. Decision-based KD focuses on a more challenging scenario where only top-1 labels are accessible.



\paragraph{Test-Time Data Augmentation.} Test-time data augmentation is a common technique in computer vision~\cite{krizhevsky2009tta, simonyan2014tta, he2016tta,test-time-aug,lyzhov2020greedytta,shanmugam2021bettertta} 
and is also feasible for natural language processing (NLP)~\cite{liu2019anonymizedbert, shleifer2019ulmfit, xu2022situ}. 
Although differing in final purpose, test-time and training-time data augmentation share a large portion of common techniques in NLP. 
Due to the discrete nature of language, one line of work conducts augmentation by modifying tokens based on rules~\cite{sahin2018aug,eda,chen2020findingaug} or models~\cite{sennrich2016bt,yang2020generativeaug,quteineh2020gpt2aug,anaby2020notaug}, and another line of work operates in embedding or representation space~\cite{chen2020mixaug,cheng2020advaug,chen2021manifoldaug,wei2022learningaug}. In this work, as we do not have access to the teacher model, we follow the first line of work to conduct test-time data augmentation. 

%% file: sections/method.tex
\section{Background}
Knowledge Distillation (KD) is a technique that aims to transfer knowledge from the teacher model to the student model by aligning certain statistics, usually the logits, of the student to those of the teacher.
Given input $x$, we denote the pre-softmax logits vector of the teacher and student as $\bm{z}$ and $\bm{v}$, respectively.
The process of KD involves minimizing the Kullback-Leibler (KL) divergence between the probabilities induced from $\bf{z}$ and $\bm{v}$ as follows:
\begin{equation}
    \mathcal{L}_{\rm KD}={\rm KL}\left({\rm softmax}(\bm{v}/\tau)||{\rm softmax}(\bm{z}/\tau)\right),
    \label{eq:loss-kd}
\end{equation}
where $\tau$ is the temperature hyper-parameter. The student model is trained by minimizing the loss function
\begin{equation}
\mathcal{L}=\mathcal{L}_{\rm CE}+\lambda\mathcal{L}_{\rm KD},
\label{eq:loss-tot}
\end{equation}
where $\mathcal{L}_{\rm CE}$ is the cross entropy loss over the ground-truth label, and $\lambda$ is the scaling factor used for balancing the importance of the two losses.

\section{Methodology}

\begin{figure*}[htpb]
    \centering
    \includegraphics[width=1.0\textwidth]{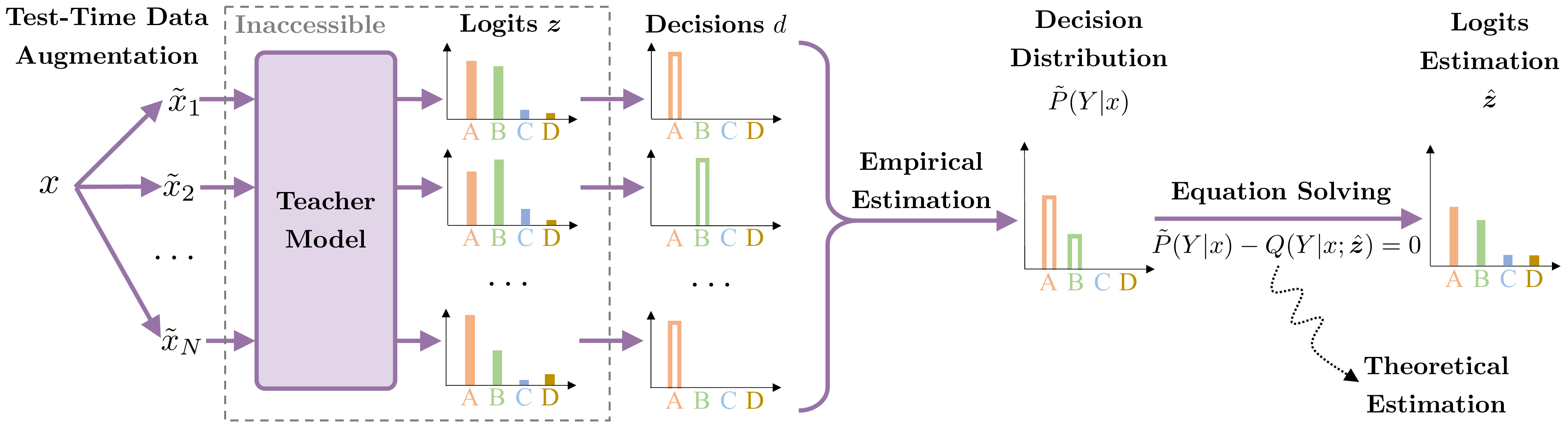}
    \caption{Overview of the proposed framework. For each training sample, the inaccessible teacher logits $\bm{z}$ contain rich information. However, the teacher model only provides decisions, which lost the information contained in most categories. We bridge this gap in two steps. First, we conduct test-time data augmentation on the original input $x$ and estimate the decision distribution $\tilde{P}(Y|x)$ with teacher decisions $d$ on the augmented data. Then, given the deicision distribution $\tilde{P}(Y|x)$, we calculate the logits estimation $\hat{\bm{z}}$ by solving the equation $\tilde{P}(Y|x)-Q(Y|x;\hat{\bm{z}})$, where $Q(Y|x;\hat{\bm{z}})$ is the theoretical estimation of the decision distribution. As a result, the information contained in original teacher logits $\bm{z}$ is partly recovered in the estimated logits $\hat{\bm{z}}$.
    }
    \label{fig:framework}
\end{figure*}



\subsection{Overview}
\label{sec:dbkd}
In the decision-based scenario, the $\bm{z}$ item in Eq.~\ref{eq:loss-kd} is not accessible. Instead, the PLM API returns the model decision $d=\mathop{\arg \max}\limits_{1\leq j\leq L}z_{j}$, where $z_j$ is the $j$-th logit in $\bm{z}$, and $L$ denotes the dimension of output label space. Obviously, $d$ only carries the information of ``the $j$-th logit is the largest''. In constrast, $\bm{z}$ contains richer information. For example, comparing $\bm{z}_1=[0.6, 0.3]$ and $\bm{z}_2=[0.9, 0.1]$, although they correpond to the same decision, i.e., $d=1$, they also imply that the second sample seems more likely to be of the first class. Therefore, there is a big information gap between logits and decsions, and our proposed method aims at narrowing down the gap by finding a better estimation of the logits.

Figure~\ref{fig:framework} shows the framework of our proposed method. The key idea is combining the empirical and theoretical estimations of the conditional decision distribution $P(Y|x)$, where $Y$ is a random variable denoting decision, to form an equation whose solution is the logits.
Specially, first we leverage test-time data augmentation to generate $N$ augmented samples for the given sample $x$ and collect the teacher decisions for them. Then, we build an empirical estimation $\tilde{P}(Y|x)$ for $P(Y|x)$ based on the decisions. Next, we derive a theoretical estimation $Q(Y|x; \hat{\bm{z}})$ parameterized by the true logits $\hat{\bm{z}}$ for $P(Y|x)$ and form the following equation 
\begin{equation}
    \tilde{P}(Y|x) - Q(Y|x; \hat{\bm{z}})=0.
    \label{eqn:root-finding}
\end{equation}
Finally, by solving for $\hat{\bm{z}}$, we get the estimated logits and the student model can be trained by conventional KD (Eq.~\ref{eq:loss-tot}).



Compared with the existing decision-based KD method~\cite{wang2021dbkd}, our method leverages data augmentation instead of binary search and optimization on the input samples. Therefore, it is applicable to discrete inputs which can hardly be searched or optimized.


\subsection{Empirical Estimation of the Conditional Decision Distribution}
\label{sec:method-detail}
In this secion, we will introduce how to get $\tilde{P}(Y|x)$ in Eq.~\ref{eqn:root-finding}. Given a sample $x$ and a teacher model $\mathcal{M}_{\theta}$ parameterized by $\theta$, we first generate $N$ augmented samples $\mathcal{X}=\{\tilde{x}_i=\mathrm{F}(x,i)\}_{i=1}^N$ with a test-time data augmentation function
$\mathrm{F}(\cdot,\cdot)$. Then the teacher decisions $\mathcal{D}=\{d_i=\mathcal{M}_\theta(\tilde{x}_i)\}$ are collected. Finally, $P(Y|x)$ is approximated as 
\begin{equation}
    \tilde{P}(Y|x) \approx \frac{1}{N}\sum_{i=1}^{N}\mathbbm{1}_{d_i},
\label{eq:emp-estimation}
\end{equation}
where $\mathbbm{1}_{d_i}\in \{0,1\}^{L}$ is a $L$-dimensional one-hot vector whose $d_i$-th element is $1$, and $L$ is the number of categories.

$\mathrm{F}(\cdot,\cdot)$ plays a crucial role in the process and an ideal $\mathrm{F}(\cdot,\cdot)$ should satisfy three requirements.
First, it should conserve true labels~\cite{test-time-aug}. Second, it should have a low computational cost, since it will be computed repetitively.
Third, the degree of noise introduced by $\mathrm{F}(\cdot,\cdot)$ should be quantizable and controllable, crucial for the following steps.
Thus, following~\citet{eda}, we define $\mathrm{F}(\cdot,\cdot)$ as an operation randomly sampled from synonym replacement, random insertion, random swap, and random deletion operations.

\input{algs/reconst.tex}

\subsection{Theoretical Estimation of the Conditional Decision Distribution}
\label{sec:recon}
In this section, we will introduce how to get the theoretical estimation $Q(Y|x; \hat{\bm{z}})$ parameterized by $\hat{\bm{z}}$ in Eq.~\ref{eqn:root-finding}. 
The outline of the derivation is that we assume the logits are sampled from an $L$-dimensional distribution $P(\bm{Z}|x)$, where $\bm{Z}=[Z_j]$ is an $L$-dimensional random variable denoting logits. Then $Q(Y=i|x; \hat{\bm{z}})$ is equal to the probability that the $i$-th dimension of $\bm{Z}$ takes the largest value, which can be calculated mathematically from $P(\bm{Z}|x)$.

Following the above outline, we have
\begin{equation}
    Q(Y=i|x;\hat{\bm{z}})=P\left(Z_i=\max_{1\leq j\leq L}Z_j\Big|x\right).
\label{eq:joint-prob}
\end{equation}
To derive the above probability, we reformulate it in terms of orthant probability.
First, we introduce an $L-1$ dimensional auxiliary random variable $\bm{U}=[U_j]$, which is defined as
\begin{equation}
    U_j =
    \begin{cases}
    Z_i-Z_j & (j<i) \\
    Z_i-Z_{j+1} & (j\geq i)
    \end{cases}
    ,1\leq j \leq L-1.
\end{equation}
Note that the $i$-th dimension of $\bm{Z}$ is eliminated due to $Z_i-Z_i=0$. Then Eq.~\ref{eq:joint-prob} can be rewritten as a non-centred orthant probability distribution
\begin{equation}
\begin{aligned}
Q(Y=i|x;\hat{\bm{z}})=P(U_j\geq 0, 1\leq j\leq L-1).
\end{aligned}
\label{eq:orthant-1}
\end{equation}

To simplify the calculation of the probability in Eq.~\ref{eq:orthant-1}, we assume $\bm{Z}$ follows a multivariate Gaussian distribution with mean $\hat{\bm{z}}$ and covariance matrix $\bm{\Sigma}$, i.e., $\bm{Z}\sim \mathcal{N}(\hat{\bm{z}},\bm{\Sigma})$. Then we have $\bm{U}\sim \mathcal{N}(\bm{\mu},\bm{R})$, where
\begin{equation}
    \mu_j =
    \begin{cases}
    \hat{z}_i-\hat{z}_j & (j<i) \\
    \hat{z}_i-\hat{z}_{j+1} & (j\geq i)
    \end{cases}
    ,1\leq j \leq L-1.
\end{equation}
And Eq.~\ref{eq:orthant-1} can be calculated by the following multiple integrations
\begin{equation}
    \int_0^{+\infty}\dots \int_0^{+\infty}\phi_{L-1}(\bm{U};\bm{\mu}, \bm{R})dU_1\dots dU_{L-1},
    \label{eq:integration}
\end{equation}
where $\phi_{L-1}(\bm{U}; \bm{\mu},\bm{R})$ is the probability density function.

We leverage the recursive algorithm proposed by \citet{orthant} to solve the above integrations. Taking $L=4$ as an example, the major steps of the algorithm are as follows. First, we decompose the covariance matrix $\bm{R}$ as $\bm{R}=\bm{BB}^T$ via Cholesky decomposition, where $\bm{B}$ is a lower triangular matrix. Then we have $\bm{U}=\bm{BM}+\bm{\mu}$,
where $\bm{M}\sim\mathcal{N}(\bm{0},\bm{I}_{L-1})$ and $\bm{I}_{L-1}$ is an identity matrix of dimension $L-1$. Next, $Q(Y=i|x;\hat{\bm{z}})$ can be further decomposed as
\begin{equation}
\label{eq:rec0}
    \begin{aligned}
    Q(Y=i&|x;\hat{\bm{z}})=P(U_j\geq0, 1\leq j\leq 3) \\
        =P(&b_{11}M_1+\mu_1\geq0,\\
            &b_{21}M_1+b_{22}M_2+\mu_2\geq0,\\
            &b_{31}M_1+b_{32}M_2+b_{33}M_3+\mu_3\geq0),
    \end{aligned}
\end{equation}
where $b_{ij}$ denotes the elements in the $\bm{B}$ matrix, $M_j$ denotes the $j$-th elements of the random variable $\bm{M}$, and $\mu_j$ denotes $j$-th the elements of $\bm{\mu}$. 
Finally, the required probability is given when $b_{ij}>0$
\begin{equation}
\label{eq:rec2}
    \begin{aligned}
    Q(Y=i|x;\hat{\bm{z}})=\int_{-\frac{\mu_1}{b_{11}}}^{+\infty} f_1(t)\phi(t)dt,
    \end{aligned}
\end{equation}
where $\phi(t)$ is the standard normal probability density function, and $f_1$ is defined as
\begin{align}
    f_1(s)&=\int_\frac{-\mu_2-b_{21}s}{b_{22}}^{+\infty} f_2(s,t)\phi(t)dt, \label{eq:rec1}\\
    f_2(s_1,s_2)&=\int_\frac{-\mu_3-b_{31}s_1-b_{32}s_2}{b_{33}}^{+\infty}\phi(t)dt\label{eq:rec3}.
\end{align}

Algorithm~\ref{alg:ouralg} summarizes the entire procedure of our proposed framework, where line~\ref{lst:line:recursive} refers to the integration steps in Eq.~\ref{eq:rec2} to \ref{eq:rec3}. We provide the proofs of the integration steps in Appendix~\ref{sec:app-proof}. In practice, we assume $\bm{\Sigma}$ is a diagonal matrix and $\Sigma_{ii}=\sigma^2$ to simplify the calculation, where $\sigma$ is a hyper-parameter of our algorithm.

%% file: algs/reconst.tex
\begin{algorithm}[!t]
\SetInd{0.1em}{2em}
   \SetKwFunction{DataAugmentation}{DataAugmentation}
    \SetKwFunction{MonteCarlo}{MonteCarlo}
    \SetKwFunction{RecursiveIntegration}{RecursiveIntegration}
    \SetKwFunction{CholeskyDecompose}{CholeskyDecompose}
    \SetKwFunction{Substitution}{Substitution}
    \caption{Teacher Logits Estimation}\label{alg:ouralg}
    \KwData{Input text $x$.}
    \KwResult{Teacher logits estimation $\hat{\bm{z}}$}
    \begin{algorithmic}[1]
    \REQUIRE
    Teacher model $\mathcal{M}_\theta$, data augmenta-\\tion transformation function $\mathrm{F}(\cdot,\cdot)$, aug-\\mented data number $N$, maximal iteration\\number $m$, error bound $\epsilon$, hyper-parameter\\$\sigma$, label number $L$.
    \\
    {\color{blue}\tcp{Empirical estimation of the conditional decision distribution}}
    \STATE $\{\tilde{x}_n\}_{n=1}^N \gets \{\mathrm{F}(x, n)\}_{n=1}^N$
    \STATE $\{d_n\}_{n=1}^N \gets \{\mathcal{M}_\theta(\tilde{x}_n)\}_{n=1}^N$
    \STATE $\tilde{P}(Y|x)\gets\frac{1}{N}\sum_{i=1}^{N}\mathbbm{1}_{d_i}$
    \\
    {\color{blue}\tcp{Solving equation to obtain $\hat{\bm{z}}$}}
    \STATE $k\gets 0$
    \STATE $\hat{\bm{z}}\gets \bm{0}$\quad{\color{blue}\tcp{$\hat{\bm{z}}=[\hat{z}_i], 1\le i \le L$}}
    \STATE Initialize $\bm{R}\in\mathbb{R}^{(L-1)\times(L-1)}$ whose diago- nal elements are $\sigma^2$ and the others are $2\sigma^2$
    \REPEAT
        \STATE $\bm{p}\gets \bm{0}$\quad{\color{blue}\tcp{$\bm{p}=[p_i], 1\le i \le L$}}
        \FOR{$i=1,i\leq L$}
        \STATE $\bm{\mu}\gets [\hat{z}_i-\hat{z}_1,\dots,\hat{z}_{i}-\hat{z}_{i-1},\hat{z}_{i}-\hat{z}_{i+1},\dots,\hat{z}_i-\hat{z}_L]$
        \STATE $\bm{B}\gets$ \CholeskyDecompose{$\bm{R}$}
        \STATE $p_i\gets$\RecursiveIntegration{$\bm{\mu},\bm{B}$}\label{lst:line:recursive}
        \STATE $i\gets i+1$
        \ENDFOR
        \STATE $k\gets k + 1$
        \STATE $\hat{\bm{z}} \gets \tilde{P}(Y|x)-\bm{p}+\hat{\bm{z}}$
    \UNTIL {$|\bm{p}-\tilde{P}(Y|x)|\leq\epsilon$ \textbf{or} $k=m$}
    \RETURN $\hat{\bm{z}}$
    \end{algorithmic}
\end{algorithm}

%% file: sections/exp.tex
\section{Experiments}


\subsection{Experimental Settings}
\paragraph{Datasets and Evaluation Metrics.} We evaluate our method on machine reading comprehension (MRC) and natural language understanding (NLU) datasets. For MRC, two widely used multiple-choice datasets RACE~\cite{race} and DREAM~\cite{dream} are used. For NLU, we select sentiment analysis dataset SST-2~\cite{sst2}, linguistic acceptability dataset CoLA~\cite{cola}, paraphrasing dataset MRPC~\cite{mrpc} and QQP~\cite{qqp}, and natural language inference (NLI) datasets RTE~\cite{rte}, MNLI~\cite{mnli}, and QNLI~\cite{qnli} as representative datasets. Following previous works~\cite{race,dream,glue}, we report Matthews correlation coefficient for CoLA, F1 and accuracy for MRPC and QQP, and accuracy for all the other datasets. For each experiment, the model is evaluated on the validation set once an epoch, and the checkpoint achieving the best validation results is evaluated on the test set. The results averaged over five random seeds are reported for the MRC datasets. Due to the submission quota, we only report results for one trial for the NLU tasks.


\paragraph{Baselines.}
We compare our method with the following four baselines:
\begin{itemize}
    \item \textit{Hard}: We regard the teacher decisions as the ground truth labels and train the student model solely with the cross entropy loss.
    \item \textit{Noisy Logits}~\cite{wang2021dbkd}: The student model is trained via the KD objective (Eq.~\ref{eq:loss-tot}) with the teacher logits replaced with randomly sampled soft labels. 
    \item \textit{Smooth}: We apply label smoothing~\cite{label-smoothing} with a smoothing factor $0.1$ on the teacher decision of the original sample, and use the smoothed decision as teacher prediction probability. This a straightforward approach to generate soft labels from teacher decisions.
\end{itemize}
To better investigate the upper bound of our method, we also leverage the following three baselines from \citet{wang2021dbkd}:
\begin{itemize}
    \item \textit{Student CE}~\cite{wang2021dbkd}: The student model is trained using only the cross entropy loss calculated from the ground-truth labels.
    \item \textit{Standard KD}: The student model is optimized with the standard KD objective (Eq.~\ref{eq:loss-tot}). Note that the teacher model is used as a white-box model in this baseline. 
    \item \textit{Surrogate}: Following~\citet{wang2021dbkd}, we train the student model via KD with a surrogate teacher, simulating training a lightweight, white-box teacher model for knowledge distillation.
\end{itemize}

\input{tables/race-main.tex}
\input{tables/dream-main.tex}
\input{tables/glue-main.tex}

\paragraph{Implementation Details.}
We implement the teacher model as the \emph{finetuned} 12-layer BERT model (BERT$_{\rm BASE}$) or 24-layer BERT model (BERT$_{\rm LARGE}$) for each task. The student model is a 4-layer or 6-layer BERT-style model.
Following previous KD works~\cite{patientkd, dynamickd}, we initialize the student model from the \emph{raw} 12-layer BERT model.
We adopt EDA~\cite{eda} as the tool for test-time data augmentation.

\subsection{Results on MRC Datasets}
Experimental results on the MRC datasets RACE and DREAM are shown in Table~\ref{tab:race-main} and Table~\ref{tab:dream-main}, respectively. In each table, we report three sets of results with different teacher and student architectures. For example, the string ``12L$\rightarrow$4L'' in the tables means we leverage the finetuned 12-layer BERT model (BERT$_{\rm BASE}$) as the teacher, and the 4-layer BERT-style model as the student.
Note that \textit{Teacher} and \textit{Standard KD} serve as the upper bounds of our method. Therefore, we do not directly compare our method with them. 
From these results, we can observe that:

(1) Our proposed method outperforms baselines consistently and significantly. The performance gap between our method and the second best baselines except \textit{Teacher} and \textit{Standard KD} are from 0.96 to 1.39 on the RACE datasets and from 0.51 to 0.72 on the DREAM dataset, indicating that narrowing down the information gap between logits and decisions are effective for decision-based KD. 

(2) Surprisingly, our proposed method achieves comparable results with \textit{Standard KD} under a few settings. \textit{Standard KD} treats the teacher as a white-box model and is an intuitive upper bound of our method. However, the smallest performance gap between our method and \textit{Standard KD} is 0.06 (12L$\rightarrow$4L on RACE-High) on the RACE datasets and is 0.25 (12L$\rightarrow$4L) on the DREAM dataset. Moreover, among all the twelve pairs of results, there is a third of them with a gap of less than 0.50. These results further justify the effectiveness of our proposed method. And we argue that this is mainly due to our better estimation of the logits.

(3) None of the baselines besides \textit{Standard KD} can consistently achieve better results than training the student model without KD (\textit{Student CE}), indicating that decision-based KD is a challenging task and the lost information from decisions compared with logits is essential. \textit{Hard} performs slightly better than \textit{Student CE} 
on the RACE datasets but significantly worse than \textit{Student CE}  on the DREAM dataset. We conjecture that this is because the teacher models have significantly better results on the RACE datasets than on the DREAM dataset, i.e., \textit{Hard} can only work well with strong teacher models, whose decisions may be less noisy and the information gap between decisions and logits is smaller. Our method can be viewed as a special form of logits smoothing. However, both \textit{Noisy Logits} and \textit{Smooth} only achieves comparable or worse results than \textit{Student CE}, indicating straightforward logits smoothing is not effective.

(4) Our method benefits from both better teacher models and larger student models. When the teacher grows larger (12L$\rightarrow$4L v.s. 24L$\rightarrow$4L), our method achieves a 1.10 performance gain on the DREAM dataset. Meanwhile, when the student models grow from 4 to 6 layers (12L$\rightarrow$4L v.s. 12L$\rightarrow$6L), the performance gains on all datasets are remarkably larger. The same trend is also observed for other baselines, suggesting that improving the capacity of the student model is a simple yet effective way to improve the performance of decision-based KD.

\subsection{Results on NLU Datasets}
Table~\ref{tab:glue-main} shows the results on the NLU datasets. First, our proposed method achieves the best results among all decision-based baselines, justifying that our method is generalizable to a large range of NLU tasks. Although our method does not outperform \textit{Surrogate} on the RTE and MNLI-mm datasets, the gap is only 0.2.
Second, the performance gap between ours and \textit{Standard KD} is also small, providing extra evidence that our method estimates the teacher logits well.
Third, all the baselines excluding \textit{Teacher} and \textit{Standard KD} have comparable performance, suggesting that decision-based KD is also challenging for NLU tasks.
Above all, in conjunction with the results on the MRC datasets, we can conclude that our method is effective for diverse tasks and model architectures. 

\subsection{Analysis on Logits Estimation}
We have conjectured that the good performance of our method comes from better logits estimation. To justify this assumption, we conduct a quantitative analysis in this section.
We compute the mean squared errors (MSEs) between the soft labels generated from each method after softmax and teacher predictions on the training set of RACE-High \footnote{As \textit{Hard} produces zero probabilities, Kullback–Leibler divergence is not applicable. A temperature ($\tau$ in Eq.~\ref{eq:loss-kd}) of 10 is leveraged when needed.}. As shown in Figure~\ref{fig:logits-eval}, the soft labels generated by our method are the closest to the teacher predictions among all methods.
However, it should be noted that the probabilities (or logits) of the teacher are not perfect, as it does not achieve perfect final performance on the dataset. Therefore, the MSEs have a positive correlation with the final performance but are not oracle indicators.

\input{figs/estimation_eval.tex}

\subsection{Ablation Study}
\begin{figure}[htpb]
    \centering
    \includegraphics[width=0.95\linewidth]{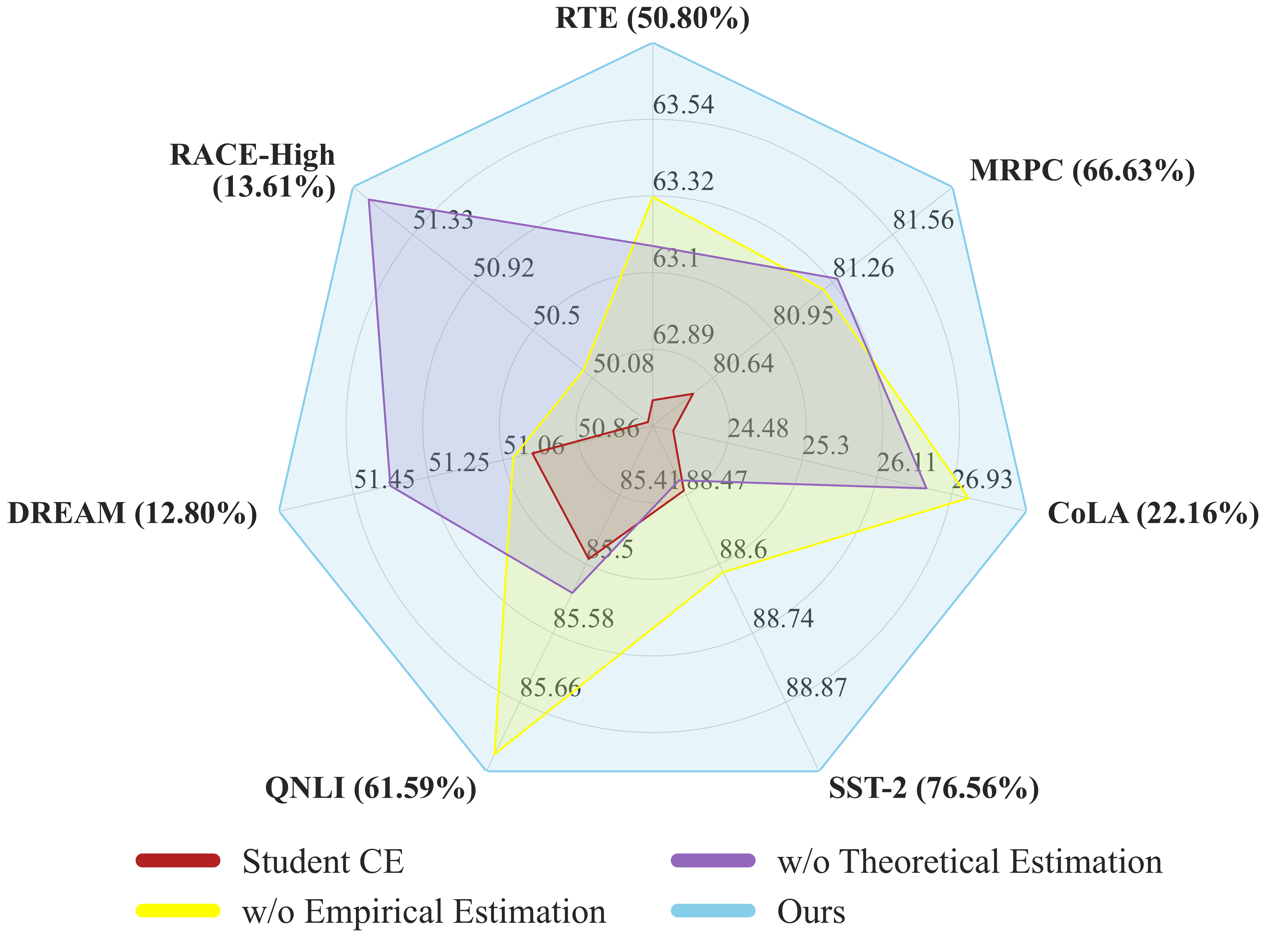}
    \caption{Ablation studies for the empirical estimation step (Section~\ref{sec:method-detail}) and the theoretical estimation step (Section~\ref{sec:recon}) on the dev set of each dataset. We also include the results from \textit{Student CE} method for comparison. The percentages after dataset names denote the ratio of empirical estimations $\tilde{P}(Y|x)$ being one-hot vectors.
    }
    \label{fig:ablation}
\end{figure}

This section consists of a series of experiments aimed at validating the contributions of different components in our method. First, we compare our method with its two variants in Figure~\ref{fig:ablation}: (1) \textit{w/o Empirical Estimations.} In this variant, we replace the $\tilde{P}(Y|x)$ in Eq.~\ref{eq:emp-estimation} with the teacher decision on original data to skip the empirical estimation step. (2) \textit{w/o Theoretical Estimation}. We replace the $Q(Y|x;\hat{\bm{z}})$ term in Eq.~\ref{eqn:root-finding} with ${\rm softmax}(\bm{z})$ to skip the theoretical estimation step. For each dataset, we also count the percentage of empirical estimations $\tilde{P}(Y|x)$ being one-hot vectors, which means that teacher decisions are consistent on augmented inputs. According to the results, we find that the performance drops for both variants on all datasets, indicating the necessity of empirical estimation and theoretical estimation. Interestingly, we also find a positive correlation between the percentage of one-hot $\tilde{P}(Y|x)$ and the performance degradation from \textit{w/o Theoretical Estimation} variant. This phenomenon highlights the capability of the theoretical estimation step to estimate teacher logits and narrow the information gap between decisions and logits even without observed decisions.

Second, we further analyze the effect of empirical estimation by changing the sampling times $N$ in Eq.~\ref{eq:emp-estimation}. As shown in Figure~\ref{fig:ablation-mc}, when $N$ increases, the performance of our method first increases and then stabilizes. Considering a larger $N$ leads to more queries to the teacher model, $N$ should be as low as possible without compromising the model performance. Therefore, $N=10$ is the optimal choice according to the results.

Finally, we investigate the contribution of Eq.~\ref{eqn:root-finding}, which combines the empirical estimation and the theoretical estimation together. In our framework, the root $\hat{\bm{z}}$ of the equation is found by fixed-point iteration, and its precision is controlled by the error bound $\epsilon$. Results in Figure~\ref{fig:ablation-eps} show a negative correlation between $\epsilon$ and KD performance. As $\epsilon$ increases from $10^{-4}$ to $10^{-1}$, the performance of our method slightly drops. When $\epsilon$ increases to $1$, which means the logits estimation becomes extremely inaccurate, the performance drops dramatically and is close to the performance of \textit{Student CE} method.

\subsection{Computational Cost Analysis}
The additional computational cost of our method compared to \textit{Standard KD} consists of two parts. The first part is test-time data augmentation, which necessitates multiple queries to the teacher model for each training sample. In this paper, we set the default number of augmented samples $N$ per training case to 10.
The second part is solving Eq.~\ref{eqn:root-finding} using the empirical estimation of decision distribution $\tilde{P}(Y|x)$, which is made negligible by pre-building a lookup table from $\tilde{P}(Y|x)$ to logits estimation $\hat{\bm z}$ before KD.
In total, the additional cost of our method mainly comes from 10 queries made to the teacher model per training sample. By contrast, the existing soft label generation method DB3KD~\cite{wang2021dbkd} requires 1,000 to 20,000 queries to the teacher model per training sample.

\input{figs/ablation-mc.tex}
\input{figs/ablation-eps.tex}

%% file: tables/race-main.tex
\begin{table*}
    \centering
    \small
    \begin{tabular}{l|c|ccc|ccc|ccc}
        \toprule
        \multirow{2}{*}{Methods} & \multirow{2}{*}{DB}&
        \multicolumn{3}{c|}{12L$\rightarrow$4L}&
        \multicolumn{3}{c|}{24L$\rightarrow$4L}&
        \multicolumn{3}{c}{12L$\rightarrow$6L} \\
        \cmidrule(lr){3-5}\cmidrule(lr){6-8}\cmidrule(lr){9-11}
        & & Middle & High & All
        & Middle & High & All
        & Middle & High & All \\
        \midrule\midrule
        Teacher & & 68.25 & 61.21 & 66.74 &
                  71.73 & 64.29 & 69.94 &
                  68.25 & 61.21 & 66.74 \\
        Standard KD & & 54.12 & 48.95 & 53.08 &
                    53.12 & 50.24 & 53.45 &                    61.69 & 53.74 & 59.28 \\
        \midrule\midrule
        Student CE & & 51.03 & 47.25 & 49.94 &
                  51.03 & 47.25 & 49.94 &
                  59.11 & 51.35 & 56.80 \\
        Surrogate & & 51.28 & 47.90 & \underline{51.03} &
                    51.28 & \underline{47.90} & \underline{51.03} &
                    \underline{59.89} & 50.81 & 56.49 \\
        Hard & \checkmark & 51.60 & \underline{47.93} & 50.29 &
                       \underline{51.59} & 46.93 & 50.78 &
                       59.64 & 51.63 & \underline{56.98} \\
        Noisy Logits & \checkmark & 50.57 & 46.30 & 49.86 &
                             50.57 & 46.30 & 49.86 &
                             58.97 & 51.12 & 55.60 \\
        Smooth & \checkmark & \underline{51.71} & 46.93 & 49.56 & 50.68 & 46.80 & 50.31 &
                                    59.68  & \underline{51.71} & 56.30 \\
        \rowcolor{Gray}
        \DBKD & \checkmark  & \textbf{52.81} & \textbf{48.89} & \textbf{52.17} & \textbf{52.98} & \textbf{49.06} & \textbf{52.08} &
                                \textbf{61.10} & \textbf{52.81} &   \textbf{58.01} \\
        \bottomrule
    \end{tabular}
    \caption{Results on the test sets of RACE. Decision-based (DB) methods are checked in the second column. BERT-style architectures are used for the teacher and student models. And the string ``12L$\rightarrow$4L'' denotes the teacher has 12 layers and the student has 4 layers. The best results besides those of \textit{Teacher} and \textit{Standard KD} are in {\bf bold}, and the \underline{second best} ones are underlined.  ``Teacher'' denotes the performance of the teacher model, respectively. The reported results are the average of five different random seeds.}
    \label{tab:race-main}
\end{table*}

%% file: tables/dream-main.tex
\begin{table}
    \centering
    \small 
    \begin{tabular}{l|c|ccc}
        \toprule
        Methods & DB & 12L$\rightarrow$4L & 24L$\rightarrow$4L & 12L$\rightarrow$6L  \\
        \midrule\midrule
        Teacher & & 60.44 & 61.69 & 60.44 \\
        Standard KD & & 51.89 & 53.37 & 54.97 \\
        \midrule\midrule
        Student CE & & 51.00 & 51.00 & 53.62 \\
        Surrogate & & 51.11 & 51.11 & 53.57 \\
        Hard & \checkmark & 50.09 & 49.98 & 51.31 \\
        Noisy Logits & \checkmark & \underline{51.13} & 51.13 & \underline{53.78} \\
        Smooth & \checkmark & 51.04 & \underline{52.03} & 53.43 \\
        \rowcolor{Gray}
        \DBKD & \checkmark & \textbf{51.64} & \textbf{52.74} & \textbf{54.50} \\
        \bottomrule
    \end{tabular}
    \caption{Results on the test set of DREAM. The reported results are the average of five different random seeds.}
    \label{tab:dream-main}
\end{table}

%% file: tables/glue-main.tex
\begin{table*}
    \centering
    \small 
    \begin{tabular}{l|c|ccccccc|c}
        \toprule
        Methods& \makecell{DB} & \makecell{RTE\\(Acc.)} & \makecell{MRPC\\(F1 / Acc.)} & \makecell{CoLA\\(Matt.)} & \makecell{QNLI\\(Acc.)} & \makecell{SST-2\\(Acc.)} & \makecell{MNLI-m / mm\\(Acc.)} & \makecell{QQP\\(F1 / Acc.)} & Average \\
        \midrule\midrule
        Teacher & & 66.2 & 87.3 / 82.3 & 53.7 & 90.9 & 93.6 & 84.4 / 83.5 & 71.4 / 89.1 & 79.1 \\
        Standard KD &  & 63.3 & 82.9 / 75.0 & 22.7 & 85.6 & 89.6 & 78.8 / 77.6 & 69.1 / 88.0 & 71.0 \\
        \midrule\midrule
        Student CE &  & 63.2 & 81.2 / 69.8 & 21.5 & 85.2 & 89.2 & \underline{78.4} / 76.7 & 67.5 / 87.3 & 69.9 \\
        Surrogate & & \textbf{63.6} & \underline{82.5} / 74.8 & 18.9 & 85.2 & \underline{89.4} & \underline{78.4} / \textbf{77.3} & 67.8 / 87.4 & 70.2 \\
        Hard & \checkmark & 63.2 & \underline{82.5} / 74.8 & 20.7 & \underline{85.6} & 89.2 & 78.1 / \underline{77.2} & \underline{68.2} / \underline{87.6} & 70.4 \\
        Noisy Logits & \checkmark & 63.3 & 81.6 / 74.0 & 21.8 & 85.3 & 88.5 & 78.1 / 76.7 & 67.5 / 87.4 & 70.2 \\
        Smooth & \checkmark & \underline{63.4} & 82.4 / \underline{75.1} & \underline{22.2} & 85.1 & 89.2 & 78.0 / 77.0 & 67.9 / 87.6 & \underline{70.6} \\
        \rowcolor{Gray}
        \DBKD & \checkmark & \underline{63.4} & \textbf{82.9} / \textbf{75.2} & \textbf{23.7} & \textbf{85.7} & \textbf{89.5} & \textbf{78.6} / 77.1 & \textbf{68.5} / \textbf{88.0} & \textbf{71.1} \\
        \bottomrule
    \end{tabular}
    \caption{Results from the GLUE test server. We show the evaluation metrics under the task names. We use BERT$_{\rm BASE}$ model as the teacher and a 4-layer BERT-style model as the student.}. 
    \label{tab:glue-main}
\end{table*}

%% file: figs/estimation_eval.tex
\begin{figure}
\centering
\small
\begin{tikzpicture}

\begin{axis} [yticklabels={\DBKD,Smooth,Surrogate,Noisy Logits,Hard},
    ytick={1,2,3,4,5},
    xtick={0, 0.03, 0.06, 0.09, 0.12, 0.15},
    x tick label style={
        /pgf/number format/.cd, fixed, fixed zerofill, precision=2, /tikz/.cd
    },
    xlabel={Mean squared error}, xlabel near ticks,
    width=0.9\linewidth,height=0.5\linewidth,
    xbar,
    bar width=7pt,
    tickwidth=0pt,
    xmin=0,xmax=0.15,
    xmajorgrids=true,
    grid style=dashed,
    nodes near coords,
    nodes near coords style={/pgf/number format/.cd,fixed,fixed zerofill,precision=4}
    ]
\addplot[xbar,fill=MyDarkThemeColor] coordinates {
	(0.0077,1) 
	(0.0100,2) 
	(0.0170,3) 
	(0.0318,4) 
	(0.1023,5)
}; 
\end{axis}
\end{tikzpicture}
\caption{Mean squared errors (MSEs) between the estimated probabilities and the ground-truth probability produced by the teacher model on the training set of RACE-High.}
\label{fig:logits-eval}
\end{figure}

%% file: figs/ablation-mc.tex
\begin{figure}
\centering
\small
\begin{tikzpicture}
\begin{axis}[
    xlabel={Sampling time $N$},
    xmin=0,
    ylabel={Accuracy}, ylabel near ticks,
    ymin=50, ymax=52,
    ytick={50.0, 50.5, 51.0, 51.5, 52.0},
    y tick label style={
        /pgf/number format/.cd, fixed, fixed zerofill, precision=2, /tikz/.cd
    },
    ymajorgrids=true,
    grid style=dashed,
    width=\linewidth,height=0.6\linewidth,
    xtick pos=left,
    ytick pos=left,
    nodes near coords,
    every node near coord/.append style={font=\tiny,color=black,xshift=24pt,yshift=0pt,anchor=east},
    coordinate style/.condition=
{\coordindex==1}{xshift=-8pt,yshift=6pt},
    coordinate style/.condition=
{\coordindex==2}{xshift=0pt,yshift=3pt},
    coordinate style/.condition=
{\coordindex==3}{xshift=-5pt,yshift=-5pt},
    coordinate style/.condition=
{\coordindex>=4}{xshift=-8pt,yshift=-6pt},
    nodes near coords style={/pgf/number format/.cd,fixed,fixed zerofill,precision=2},
]

\addplot[
    color=MyLineChartColor,
    mark=o,
    line width=1.5pt,
    ]
    coordinates {
    (1,50.15)
    (2,50.39)
    (5,50.46)
    (10,51.75)
    (20,51.77)
    (40,51.54)
    };
    
\end{axis}
\end{tikzpicture}
\caption{Comparison of different sampling times $N$ of test-time data augmentation in the empirical estimation step. Averaged results on the dev set of RACE-High over 5 different random seeds are reported.}
\label{fig:ablation-mc}
\end{figure}
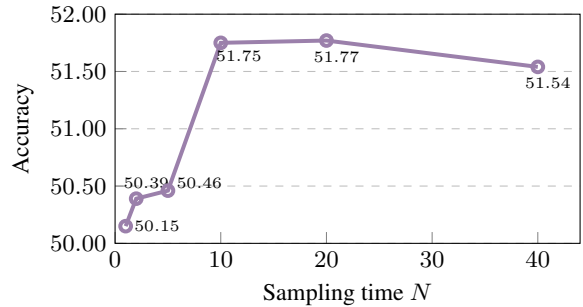

%% file: figs/ablation-eps.tex
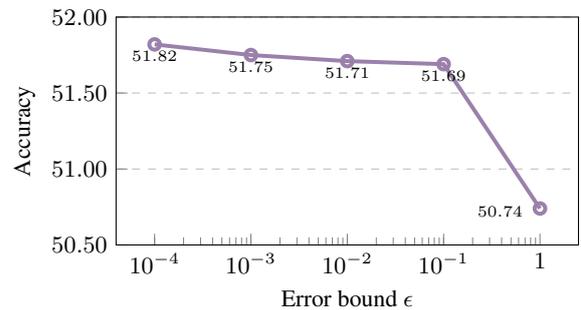
\begin{figure}
\centering
\small
\resizebox{\columnwidth}{!}{
\begin{tikzpicture}
\begin{semilogxaxis}[
    xlabel={Error bound $\epsilon$},
    ylabel={Accuracy}, ylabel near ticks,
    ymin=50.5, ymax=52,
    xmode=log,
    xtick={1e-4, 1e-3, 1e-2, 1e-1, 1},
    xticklabels={$10^{-4}$,$10^{-3}$,$10^{-2}$,$10^{-1}$,$1$},
    ytick={50.5, 51.00, 51.50, 52.00},
    y tick label style={
        /pgf/number format/.cd, fixed, fixed zerofill, precision=2, /tikz/.cd
    },
    ymajorgrids=true,
    grid style=dashed,
    width=\linewidth,height=0.6\linewidth,
    xtick pos=left,
    ytick pos=left,
    nodes near coords,
    every node near coord/.append style={font=\tiny,color=black,yshift=-10pt},
    coordinate style/.condition=
{\coordindex<1}{xshift=-15pt,yshift=3pt},
    nodes near coords style={/pgf/number format/.cd,fixed,fixed zerofill,precision=2}
]

\addplot[
    color=MyLineChartColor,
    mark=o,
    line width=1.5pt,
    ]
    coordinates {
    (1,50.74)
    (1e-1,51.69)
    (1e-2,51.71)
    (1e-3,51.75)
    (1e-4,51.82)
    };
    
\end{semilogxaxis}
\end{tikzpicture}
}
\caption{Comparison of different error bounds $\epsilon$ of teacher logits estimation in the theoretical derivation step. Averaged results on the dev set of RACE-High over 5 different random seeds are reported.}
\label{fig:ablation-eps}
\end{figure}

%% file: sections/conclusion.tex
\section{Conclusion}
We introduce a novel decision-based KD method, which bridges the information gap between teacher decisions and logits by estimating teacher logits. In contrast to existing solutions for decision-based KD, our method is applicable to NLP tasks with discrete inputs. Extensive experiments over various tasks and model architectures demonstrate the effectiveness of our proposed method.

One future direction for the decision-based KD is the exploration of other NLP tasks, such as neural machine translation, text generation, and question answering. The other direction is KD from non-NN models to NN models, which benefits the training of NN models with additional information from a wider range of models. Unlike conventional KD, decision-based KD does not require internal information from NN models and is promising for solving this problem.

%% file: sections/limitations.tex
\section*{Limitations}
This study has two main limitations. The first limitation is its reliance on the assumption that teacher logits on augmented data follow a Gaussian distribution. This assumption is used in the derivation of teacher logits in Section~\ref{sec:recon}. However, in practice, teacher logits may not strictly follow a Gaussian distribution. It is challenging to estimate teacher logits under more realistic assumptions, which requires thorough investigations on the distribution of teacher logits and more complex computations for logits estimation.

The second limitation is that our method still requires access to the training dataset of the downstream tasks. In this paper, we focus on KD when teacher PLMs only return decisions. However, our method is not capable of KD without publicly available training data, which is a more challenging scenario for decision-based KD. We believe training a data generation model~\cite{wang2021dbkd, qekd, hardstealing} might be useful for such cases.

%% file: sections/ethics.tex
\section*{Ethics Statement}
In ethical considerations, our method risks being used as a means of model stealing. Therefore, defensive techniques against the proposed method are required. However, it also has significant positive implications. On the one hand, it can serve as a powerful tool for research on model extraction attacks, thereby promoting the advancement of related studies. On the other hand, it has practical applications in real-world scenarios. For instance, a company may prefer to use a smaller model due to cost considerations, and our method allows for the easy distillation of smaller models without requiring white-box access to larger models. Additionally, our method can be used to distill non-NN models into NN models, reducing the number of model types that need to be maintained and simplifying operation and maintenance.

%% file: sections/acknowledgement.tex
\section*{Acknowledgement}
This work is supported by the National Key R\&D Program of China (2022ZD0160502) and the National Natural Science Foundation of China (No. 61925601, 62276152, 62236011). We thank all anonymous reviewers for their valuable comments and suggestions
on this work. We also thank Shuo Wang and Xiaoyue Mi for their suggestions on the writing.

%% file: sections/appendix.tex
\appendix

\section{Appendix}
\label{sec:appendix}
\subsection{Experiment Details}
\paragraph{Dataset Details}
In this paper, we use seven different datasets, and all of them are in the English language. We downloaded these datasets from the Datasets~\cite{lhoest2021datasets} library of version 2.4.0, and our use is consistent with their intended use. The other details of the datasets we used are summarized in Table~\ref{tab:dataset-info}.
\input{tables/datasets-info.tex}
\paragraph{Model Details}
We used BERT-like models~\cite{bert} in our experiments, including BERT$_{\rm BASE}$ (110M parameters), BERT$_{\rm LARGE}$ (340M parameters), 4-layer BERT-like models (53M parameters), and 6-layer BERT-like models (67M parameters). For BERT$_{\rm BASE}$ and BERT$_{\rm LARGE}$, the raw model checkpoints are obtained from Huggingface Transformers~\cite{huggingface} platform. Following \citet{dynamickd}, we initialize the 4-layer and 6-layer BERT-like models from the first 4 and 6 layers of the raw BERT$_{\rm BASE}$ model, respectively.
\paragraph{Other Details}
We finetune the BERT$_{\rm BASE}$ and BERT$_{\rm LARGE}$ models for 4 epochs. Following~\citet{cascadebert}, we train small 4-layer or 6-layer models for 10 epochs.
We use a learning rate of $5\times10^{-5}$ for MRC tasks and $2\times10^{-5}$ for NLU tasks. $\sigma$ in Algorithm~\ref{alg:ouralg} is tuned from $\{0.5, 1, 2, 4\}$, $\lambda$ in Eq.~\ref{eq:loss-tot} is tuned from $\{0.2, 0.5, 0.7\}$, and $\tau$ in Eq.~\ref{eq:loss-kd} is tuned from $\{5,10,20\}$ expect for our method, which performs better in the range of $\{1,2,4\}$.
The $\alpha$ parameters of all operations in EDA are sampled from a half-normal distribution, and we adjust the scale of the distribution to align its expectation with the default $\alpha=0.1$ in EDA.
For MRC tasks, following \citet{patientkd}, we concatenate the input passage and the question with a $\rm [SEP]$ token and append each answer at the end of the question.
The random seeds we used for experiments on MRC datasets and ablation studies on NLU datasets are from 1 to 5. The random seed for teacher training and other NLU experiments is 1. The training of a 4-layer student model on one RTX 3090 Ti GPU costs approximately 6.5 hours for our method.

\subsection{Experimental Results on Generative Language Models}
Theoretically, our method can be applied to generative LMs. In this paper, we evaluate the effectiveness of our method on the RACE dataset. We finetune a 12-layer GPT-2~\cite{gpt2} teacher to predict the class label (A/B/C/D) given the context, question, and options as prompt. Then we distill the teacher model to a 4-layer GPT-2 student. For the student model, the output vocabulary at the answer position is restricted to class label tokens. Table~\ref{tab:gpt} shows the performance of our method and baseline methods on the dev set of RACE-High. Our method significantly outperforms decision-based baseline methods and the student CE method.

\input{tables/gpt}

Different from classification tasks, our method will require much more queries to the teacher model on generation tasks because of the following reasons:
\begin{enumerate}
    \item The label space dimension $L$ in generative tasks is equal to the vocabulary size, which is quite large. The computational cost of building the look-up table will increase.
    \item For each position $i$ in a sequence, our method estimates the logits in the $i$-th position given all the tokens before the $i$-th position. Therefore, to estimate the logits in the entire sequence, we need to sample teacher decisions in each position. As a result, the computational cost will multiply by the sequence length.
\end{enumerate}

Therefore, how to improve the efficiency of applying our method to generation tasks is an interesting future research direction.

\subsection{Detailed Analysis of the Data Augmentation Techniques}
In this paper, we use the EDA augmentation tool~\cite{eda} for each sample, including its $\alpha$ parameter and the following four default augmentation techniques. Given a input sentence, a $\alpha$ parameter, and the sentence length $l_{sent}$, the four techniques can be describe as following:
\begin{enumerate}
    \item \textbf{Synonym Replacement}: First, select $\alpha l_{sent}$ words that are not stop words randomly. Second, replace each word with a random WordNet~\cite{wordnet} synonym of itself.
    \item \textbf{Random Insertion}: First, select a word in the sentence that is not a stop word randomly. Second, find a random synonym of the selected word. Third, insert the synonym into a random position in the sentence. Finally, do the above steps $\alpha l_{sent}$ times.
    \item \textbf{Random Swap}: First, choose two words in the sentence randomly. Second, swap the positions of the chosen words. Finally, do the above steps $\alpha l_{sent}$ times.
    \item \textbf{Random Deletion}: Remove each word in the sentence with probability $\alpha$ randomly.
\end{enumerate}

In Table~\ref{tab:ablation-eda}, we provide further ablation studies on each technique of EDA. We also include the performance of \textit{Surrogate} method which is the best decision-based baseline. According to the results, our method is robust to different data augmentation techniques.

\input{tables/ablation-eda}

\subsection{Proofs}
\label{sec:app-proof}
In this section, we provide the detailed proofs of Eq.~\ref{eq:rec2} to \ref{eq:rec3}.

In Section~\ref{sec:recon}, we assume $\Sigma$ is a diagonal matrix and $\Sigma_{ii}=\sigma^2$. Therefore, the diagonal elements of $\bm{B}$ are positive, and Eq.~\ref{eq:rec0} can be rewritten as:
\begin{equation}
\begin{aligned}
    Q(Y=i|x;\hat{z})=P(&M_1 \geq \frac{-\mu_1}{b_{11}},\\
    M_2 \geq& \frac{-\mu_2-b_{21}M_1}{b_{22}},\\
    M_3 \geq& \frac{-\mu_3-b_{31}M_1-b_{32}M_2}{b_{33}}).
\end{aligned}
\end{equation}

Then $Q(Y=i|x;\hat{z})$ can be calculated recursively. Eq.~\ref{eq:rec2} is an integration on $M_1 \geq \frac{-\mu_1}{b_{11}}$,
while Eq.~\ref{eq:rec1} and Eq.~\ref{eq:rec3} integrate on $M_2 \geq \frac{-\mu_2-b_{21}M_1}{b_{22}}$ and $M_3 \geq \frac{-\mu_3-b_{31}M_1-b_{32}M_2}{b_{33}}$, respectively.

%% file: tables/datasets-info.tex
\begin{table*}[t]
    \small 
    \centering
    \begin{tabular}{l|c|c|c}
         \toprule
         Name & Number of train / dev / test & License & Domain \\
         \midrule
         RACE-All~\cite{race} & 87,866 / 4,887 / 4,934 & \multirow{3}{*}{unknown} & \multirow{3}{*}{examinations} \\
         RACE-Middle~\cite{race} & 25,421 / 1,436 / 1,436 & & \\
         RACE-High~\cite{race} & 62,445 / 3,451 / 3,498 & & \\
         \midrule
         DREAM~\cite{dream} & 6,116 / 2,040 / 2,041 & unknown & dialogue \\
         \midrule
         RTE~\cite{rte} & 2,490 / 277 / 3,000 & \multirow{5}{*}{CC-BY-4.0} & news, Wikipedia \\
         \cmidrule{1-2}\cmidrule{4-4}
         MRPC~\cite{mrpc} & 3,668 / 408 / 1,725 & & news \\
         \cmidrule{1-2}\cmidrule{4-4}
         CoLA~\cite{cola} & 8,551 / 1,043 / 1,063 & & misc. \\
         \cmidrule{1-2}\cmidrule{4-4}
         SST-2~\cite{sst2} & 67,349 / 872 / 1,821 & & movie reviews \\
         \cmidrule{1-2}\cmidrule{4-4}
         QNLI~\cite{qnli} & 104,743 / 5,463 / 5,463 & & Wikipedia \\
         \bottomrule
    \end{tabular}
    \caption{The detailed information of datasets we used in this paper.}
    \label{tab:dataset-info}
\end{table*}

%% file: tables/gpt.tex
\begin{table}
    \centering
    \begin{tabular}{l|c}
        \toprule
        Methods & RACE-High \\
        \midrule
        Student CE & 39.79 \\
        \midrule
        Noisy Logits & 30.19 \\
        Surrogate & 37.86 \\
        Smooth & 41.59 \\
        Hard & 42.92 \\
        \rowcolor{Gray}
        \DBKD & \textbf{44.13} \\
        \bottomrule
    \end{tabular}
    \caption{Averaged results of generative models over 5 different random seeds. All methods are evaluated on the RACE-High dev set.}
    \label{tab:gpt}
\end{table}

%% file: tables/ablation-eda.tex
\begin{table}[htpb]
    \centering
    \begin{tabular}{l|c}
        \toprule
        Methods & RACE-High \\
        \midrule
        \rowcolor{Gray}
        \DBKD & \textbf{51.75} \\
        w/o synonym replacement & 51.69 \\
        w/o random insertion & 51.44 \\
        w/o random swap & \textbf{51.75} \\
        w/o random deletion & 51.60 \\
        Surrogate (best baseline) & 50.33 \\
        \bottomrule
    \end{tabular}
    \caption{Averaged results of ablation methods over 5 different random seeds. For each ablation, we remove one of the four data augmentation techniques from EDA and evaluate it on the RACE-High dev set.}
    \label{tab:ablation-eda}
\end{table}

%% file: acl2023.bbl
\begin{thebibliography}{62}
\expandafter\ifx\csname natexlab\endcsname\relax\def\natexlab#1{#1}\fi

\bibitem[{Anaby-Tavor et~al.(2020)Anaby-Tavor, Carmeli, Goldbraich, Kantor,
  Kour, Shlomov, Tepper, and Zwerdling}]{anaby2020notaug}
Ateret Anaby-Tavor, Boaz Carmeli, Esther Goldbraich, Amir Kantor, George Kour,
  Segev Shlomov, Naama Tepper, and Naama Zwerdling. 2020.
\newblock Do not have enough data? deep learning to the rescue!
\newblock In \emph{AAAI 2020}.

\bibitem[{Ayhan and Berens(2018)}]{ayhan2018tta}
Murat~Seckin Ayhan and Philipp Berens. 2018.
\newblock Test-time data augmentation for estimation of heteroscedastic
  aleatoric uncertainty in deep neural networks.
\newblock In \emph{MIDL 2018}.

\bibitem[{Bentivogli et~al.(2009)Bentivogli, Clark, Dagan, and
  Giampiccolo}]{rte}
Luisa Bentivogli, Peter Clark, Ido Dagan, and Danilo Giampiccolo. 2009.
\newblock The fifth {PASCAL} recognizing textual entailment challenge.
\newblock In \emph{TAC 2009}.

\bibitem[{Brown et~al.(2020)Brown, Mann, Ryder, Subbiah, Kaplan, Dhariwal,
  Neelakantan, Shyam, Sastry, Askell, Agarwal, Herbert-Voss, Krueger, Henighan,
  Child, Ramesh, Ziegler, Wu, Winter, Hesse, Chen, Sigler, Litwin, Gray, Chess,
  Clark, Berner, McCandlish, Radford, Sutskever, and Amodei}]{gpt3}
Tom Brown, Benjamin Mann, Nick Ryder, Melanie Subbiah, Jared~D Kaplan, Prafulla
  Dhariwal, Arvind Neelakantan, Pranav Shyam, Girish Sastry, Amanda Askell,
  Sandhini Agarwal, Ariel Herbert-Voss, Gretchen Krueger, Tom Henighan, Rewon
  Child, Aditya Ramesh, Daniel Ziegler, Jeffrey Wu, Clemens Winter, Chris
  Hesse, Mark Chen, Eric Sigler, Mateusz Litwin, Scott Gray, Benjamin Chess,
  Jack Clark, Christopher Berner, Sam McCandlish, Alec Radford, Ilya Sutskever,
  and Dario Amodei. 2020.
\newblock Language models are few-shot learners.
\newblock In \emph{NeurIPS 2020}.

\bibitem[{Chen et~al.(2021)Chen, Fan, Zhang, Chen, and
  Huang}]{chen2021manifoldaug}
Guandan Chen, Kai Fan, Kaibo Zhang, Boxing Chen, and Zhongqiang Huang. 2021.
\newblock Manifold adversarial augmentation for neural machine translation.
\newblock In \emph{Findings of the ACL 2021}.

\bibitem[{Chen et~al.(2020{\natexlab{a}})Chen, Ji, and
  Evans}]{chen2020findingaug}
Hannah Chen, Yangfeng Ji, and David Evans. 2020{\natexlab{a}}.
\newblock Finding {F}riends and flipping frenemies: Automatic paraphrase
  dataset augmentation using graph theory.
\newblock In \emph{Findings of the EMNLP 2020}.

\bibitem[{Chen et~al.(2020{\natexlab{b}})Chen, Yang, and Yang}]{chen2020mixaug}
Jiaao Chen, Zichao Yang, and Diyi Yang. 2020{\natexlab{b}}.
\newblock {M}ix{T}ext: Linguistically-informed interpolation of hidden space
  for semi-supervised text classification.
\newblock In \emph{ACL 2020}.

\bibitem[{Chen et~al.(2017)Chen, Zhang, Zhang, and Zhao}]{qqp}
Zihang Chen, Hongbo Zhang, Xiaoji Zhang, and Leqi Zhao. 2017.
\newblock Quora question pairs.

\bibitem[{Cheng et~al.(2020)Cheng, Jiang, Macherey, and
  Eisenstein}]{cheng2020advaug}
Yong Cheng, Lu~Jiang, Wolfgang Macherey, and Jacob Eisenstein. 2020.
\newblock {A}dv{A}ug: Robust adversarial augmentation for neural machine
  translation.
\newblock In \emph{ACL 2020}.

\bibitem[{Devlin et~al.(2019)Devlin, Chang, Lee, and Toutanova}]{bert}
Jacob Devlin, Ming-Wei Chang, Kenton Lee, and Kristina Toutanova. 2019.
\newblock {BERT}: Pre-training of deep bidirectional transformers for language
  understanding.
\newblock In \emph{NAACL 2019}.

\bibitem[{Dolan and Brockett(2005)}]{mrpc}
Bill Dolan and Chris Brockett. 2005.
\newblock Automatically constructing a corpus of sentential paraphrases.
\newblock In \emph{IWP 2005}.

\bibitem[{He et~al.(2016)He, Zhang, Ren, and Sun}]{he2016tta}
Kaiming He, Xiangyu Zhang, Shaoqing Ren, and Jian Sun. 2016.
\newblock Deep residual learning for image recognition.
\newblock In \emph{CVPR 2016}.

\bibitem[{Hinton et~al.(2015)Hinton, Vinyals, and Dean}]{hintonkd}
Geoffrey Hinton, Oriol Vinyals, and Jeff Dean. 2015.
\newblock Distilling the knowledge in a neural network.

\bibitem[{Jiao et~al.(2020)Jiao, Yin, Shang, Jiang, Chen, Li, Wang, and
  Liu}]{tinybert}
Xiaoqi Jiao, Yichun Yin, Lifeng Shang, Xin Jiang, Xiao Chen, Linlin Li, Fang
  Wang, and Qun Liu. 2020.
\newblock Tiny{BERT}: Distilling {BERT} for natural language understanding.
\newblock In \emph{Findings of EMNLP 2020}.

\bibitem[{Kariyappa et~al.(2021)Kariyappa, Prakash, and
  Qureshi}]{Kariyappa_2021_CVPR}
Sanjay Kariyappa, Atul Prakash, and Moinuddin~K Qureshi. 2021.
\newblock {MAZE}: Data-free model stealing attack using zeroth-order gradient
  estimation.
\newblock In \emph{CVPR 2021}.

\bibitem[{Krizhevsky et~al.(2009)Krizhevsky, Hinton et~al.}]{krizhevsky2009tta}
Alex Krizhevsky, Geoffrey Hinton, et~al. 2009.
\newblock Learning multiple layers of features from tiny images.
\newblock Technical report, University of Toronto.

\bibitem[{Lai et~al.(2017)Lai, Xie, Liu, Yang, and Hovy}]{race}
Guokun Lai, Qizhe Xie, Hanxiao Liu, Yiming Yang, and Eduard Hovy. 2017.
\newblock {RACE}: Large-scale reading comprehension dataset from examinations.
\newblock In \emph{EMNLP 2017}.

\bibitem[{Lhoest et~al.(2021)Lhoest, del Moral, Jernite, Thakur, von Platen,
  Patil, Chaumond, Drame, Plu, Tunstall et~al.}]{lhoest2021datasets}
Quentin Lhoest, Albert~Villanova del Moral, Yacine Jernite, Abhishek Thakur,
  Patrick von Platen, Suraj Patil, Julien Chaumond, Mariama Drame, Julien Plu,
  Lewis Tunstall, et~al. 2021.
\newblock Datasets: A community library for natural language processing.
\newblock In \emph{EMNLP 2021: System Demonstrations}.

\bibitem[{Li et~al.(2020)Li, Liu, Zhao, Xu, Yang, and Jin}]{emdkd}
Jianquan Li, Xiaokang Liu, Honghong Zhao, Ruifeng Xu, Min Yang, and Yaohong
  Jin. 2020.
\newblock B{ERT}-{EMD}: Many-to-many layer mapping for {BERT} compression with
  earth mover’s distance.
\newblock In \emph{EMNLP 2020}.

\bibitem[{Li et~al.(2021{\natexlab{a}})Li, Lin, Chen, Ren, Li, Zhou, and
  Sun}]{cascadebert}
Lei Li, Yankai Lin, Deli Chen, Shuhuai Ren, Peng Li, Jie Zhou, and Xu~Sun.
  2021{\natexlab{a}}.
\newblock Cascade{BERT}: Accelerating inference of pre-trained language models
  via calibrated complete models cascade.
\newblock In \emph{Findings of EMNLP 2021}.

\bibitem[{Li et~al.(2021{\natexlab{b}})Li, Lin, Ren, Li, Zhou, and
  Sun}]{dynamickd}
Lei Li, Yankai Lin, Shuhuai Ren, Peng Li, Jie Zhou, and Xu~Sun.
  2021{\natexlab{b}}.
\newblock Dynamic knowledge distillation for pre-trained language models.
\newblock In \emph{EMNLP 2021}.

\bibitem[{Liu(2019)}]{liu2019anonymizedbert}
Bo~Liu. 2019.
\newblock Anonymized {BERT}: An augmentation approach to the gendered pronoun
  resolution challenge.
\newblock In \emph{Proceedings of the First Workshop on Gender Bias in Natural
  Language Processing}.

\bibitem[{Liu et~al.(2020)Liu, Zhou, Wang, Zhao, Deng, and Ju}]{fastbert}
Weijie Liu, Peng Zhou, Zhiruo Wang, Zhe Zhao, Haotang Deng, and Qi~Ju. 2020.
\newblock Fast{BERT}: a self-distilling {BERT} with adaptive inference time.
\newblock In \emph{ACL 2020}.

\bibitem[{Liu et~al.(2019)Liu, Ott, Goyal, Du, Joshi, Chen, Levy, Lewis,
  Zettlemoyer, and Stoyanov}]{roberta}
Yinhan Liu, Myle Ott, Naman Goyal, Jingfei Du, Mandar Joshi, Danqi Chen, Omer
  Levy, Mike Lewis, Luke Zettlemoyer, and Veselin Stoyanov. 2019.
\newblock Ro{BERT}a: A robustly optimized {BERT} pretraining approach.

\bibitem[{Lyzhov et~al.(2020)Lyzhov, Molchanova, Ashukha, Molchanov, and
  Vetrov}]{lyzhov2020greedytta}
Alexander Lyzhov, Yuliya Molchanova, Arsenii Ashukha, Dmitry Molchanov, and
  Dmitry Vetrov. 2020.
\newblock Greedy policy search: A simple baseline for learnable test-time
  augmentation.
\newblock In \emph{Conference on Uncertainty in Artificial Intelligence}. PMLR.

\bibitem[{Miller(1995)}]{wordnet}
George~A. Miller. 1995.
\newblock Wordnet: A lexical database for english.
\newblock \emph{Commun. ACM}.

\bibitem[{Miwa et~al.(2003)Miwa, Hayter, and Kuriki}]{orthant}
Tetsuhisa Miwa, AJ~Hayter, and Satoshi Kuriki. 2003.
\newblock The evaluation of general non-centred orthant probabilities.
\newblock \emph{Journal of the Royal Statistical Society: Series B (Statistical
  Methodology)}.

\bibitem[{Orekondy et~al.(2019)Orekondy, Schiele, and
  Fritz}]{Orekondy_2019_CVPR}
Tribhuvanesh Orekondy, Bernt Schiele, and Mario Fritz. 2019.
\newblock Knockoff nets: Stealing functionality of black-box models.
\newblock In \emph{CVPR 2019}.

\bibitem[{Ouyang et~al.(2022)Ouyang, Wu, Jiang, Almeida, Wainwright, Mishkin,
  Zhang, Agarwal, Slama, Ray et~al.}]{instructgpt}
Long Ouyang, Jeff Wu, Xu~Jiang, Diogo Almeida, Carroll~L Wainwright, Pamela
  Mishkin, Chong Zhang, Sandhini Agarwal, Katarina Slama, Alex Ray, et~al.
  2022.
\newblock Training language models to follow instructions with human feedback.

\bibitem[{Quteineh et~al.(2020)Quteineh, Samothrakis, and
  Sutcliffe}]{quteineh2020gpt2aug}
Husam Quteineh, Spyridon Samothrakis, and Richard Sutcliffe. 2020.
\newblock Textual data augmentation for efficient active learning on tiny
  datasets.
\newblock In \emph{EMNLP 2020}.

\bibitem[{Radford et~al.(2019)Radford, Wu, Child, Luan, Amodei, Sutskever
  et~al.}]{gpt2}
Alec Radford, Jeffrey Wu, Rewon Child, David Luan, Dario Amodei, Ilya
  Sutskever, et~al. 2019.
\newblock Language models are unsupervised multitask learners.

\bibitem[{Raffel et~al.(2020)Raffel, Shazeer, Roberts, Lee, Narang, Matena,
  Zhou, Li, Liu et~al.}]{t5}
Colin Raffel, Noam Shazeer, Adam Roberts, Katherine Lee, Sharan Narang, Michael
  Matena, Yanqi Zhou, Wei Li, Peter~J Liu, et~al. 2020.
\newblock Exploring the limits of transfer learning with a unified text-to-text
  transformer.
\newblock \emph{JMLR}.

\bibitem[{Rajpurkar et~al.(2016)Rajpurkar, Zhang, Lopyrev, and Liang}]{qnli}
Pranav Rajpurkar, Jian Zhang, Konstantin Lopyrev, and Percy Liang. 2016.
\newblock S{Q}u{AD}: 100, 000+ questions for machine comprehension of text.
\newblock In \emph{EMNLP 2016}.

\bibitem[{{\c{S}}ahin and Steedman(2018)}]{sahin2018aug}
G{\"o}zde~G{\"u}l {\c{S}}ahin and Mark Steedman. 2018.
\newblock Data augmentation via dependency tree morphing for low-resource
  languages.
\newblock In \emph{EMNLP 2018}.

\bibitem[{Sanh et~al.(2019)Sanh, Debut, Chaumond, and Wolf}]{distilbert}
Victor Sanh, Lysandre Debut, Julien Chaumond, and Thomas Wolf. 2019.
\newblock Distil{BERT}, a distilled version of {BERT}: smaller, faster, cheaper
  and lighter.
\newblock In \emph{NeurIPS 2019 Workshop on Energy Efficient Machine Learning
  and Cognitive Computing}.

\bibitem[{Sanyal et~al.(2022)Sanyal, Addepalli, and Babu}]{hardstealing}
Sunandini Sanyal, Sravanti Addepalli, and R~Venkatesh Babu. 2022.
\newblock Towards data-free model stealing in a hard label setting.
\newblock In \emph{CVPR 2022}.

\bibitem[{Sennrich et~al.(2016)Sennrich, Haddow, and Birch}]{sennrich2016bt}
Rico Sennrich, Barry Haddow, and Alexandra Birch. 2016.
\newblock Improving neural machine translation models with monolingual data.
\newblock In \emph{ACL 2016}.

\bibitem[{Shanmugam et~al.(2021)Shanmugam, Blalock, Balakrishnan, and
  Guttag}]{shanmugam2021bettertta}
Divya Shanmugam, Davis Blalock, Guha Balakrishnan, and John Guttag. 2021.
\newblock Better aggregation in test-time augmentation.
\newblock In \emph{ICCV 2021}.

\bibitem[{Shleifer(2019)}]{shleifer2019ulmfit}
Sam Shleifer. 2019.
\newblock Low resource text classification with ulmfit and backtranslation.

\bibitem[{Simonyan and Zisserman(2015)}]{simonyan2014tta}
Karen Simonyan and Andrew Zisserman. 2015.
\newblock Very deep convolutional networks for large-scale image recognition.
\newblock In \emph{ICLR 2015}.

\bibitem[{Smith and Gal(2018)}]{smith2018tta}
Lewis Smith and Yarin Gal. 2018.
\newblock Understanding measures of uncertainty for adversarial example
  detection.

\bibitem[{Socher et~al.(2013)Socher, Perelygin, Wu, Chuang, Manning, Ng, and
  Potts}]{sst2}
Richard Socher, Alex Perelygin, Jean Wu, Jason Chuang, Christopher~D Manning,
  Andrew~Y Ng, and Christopher Potts. 2013.
\newblock Recursive deep models for semantic compositionality over a sentiment
  treebank.
\newblock In \emph{EMNLP 2013}.

\bibitem[{Sun et~al.(2019{\natexlab{a}})Sun, Yu, Chen, Yu, Choi, and
  Cardie}]{dream}
Kai Sun, Dian Yu, Jianshu Chen, Dong Yu, Yejin Choi, and Claire Cardie.
  2019{\natexlab{a}}.
\newblock Dream: A challenge data set and models for dialogue-based reading
  comprehension.
\newblock \emph{TACL 2019}.

\bibitem[{Sun et~al.(2019{\natexlab{b}})Sun, Cheng, Gan, and Liu}]{patientkd}
Siqi Sun, Yu~Cheng, Zhe Gan, and Jingjing Liu. 2019{\natexlab{b}}.
\newblock Patient knowledge distillation for bert model compression.

\bibitem[{Szegedy et~al.(2016)Szegedy, Vanhoucke, Ioffe, Shlens, and
  Wojna}]{label-smoothing}
Christian Szegedy, Vincent Vanhoucke, Sergey Ioffe, Jon Shlens, and Zbigniew
  Wojna. 2016.
\newblock Rethinking the inception architecture for computer vision.
\newblock In \emph{CVPR 2016}.

\bibitem[{Tang et~al.(2019)Tang, Lu, Liu, Mou, Vechtomova, and
  Lin}]{tang2019distilling}
Raphael Tang, Yao Lu, Linqing Liu, Lili Mou, Olga Vechtomova, and Jimmy Lin.
  2019.
\newblock Distilling task-specific knowledge from bert into simple neural
  networks.

\bibitem[{Truong et~al.(2021)Truong, Maini, Walls, and
  Papernot}]{Truong_2021_CVPR}
Jean-Baptiste Truong, Pratyush Maini, Robert~J. Walls, and Nicolas Papernot.
  2021.
\newblock Data-free model extraction.
\newblock In \emph{CVPR 2021}.

\bibitem[{Wang et~al.(2018)Wang, Singh, Michael, Hill, Levy, and Bowman}]{glue}
Alex Wang, Amanpreet Singh, Julian Michael, Felix Hill, Omer Levy, and Samuel
  Bowman. 2018.
\newblock {GLUE}: A multi-task benchmark and analysis platform for natural
  language understanding.
\newblock In \emph{EMNLP 2018 Workshop BlackboxNLP: Analyzing and Interpreting
  Neural Networks for NLP}.

\bibitem[{Wang et~al.(2020)Wang, Li, Wang, and Gong}]{Wang_2020_CVPR}
Dongdong Wang, Yandong Li, Liqiang Wang, and Boqing Gong. 2020.
\newblock Neural networks are more productive teachers than human raters:
  Active mixup for data-efficient knowledge distillation from a blackbox model.
\newblock In \emph{CVPR 2020}.

\bibitem[{Wang et~al.(2019)Wang, Li, Aertsen, Deprest, Ourselin, and
  Vercauteren}]{test-time-aug}
Guotai Wang, Wenqi Li, Michael Aertsen, Jan Deprest, S{\'e}bastien Ourselin,
  and Tom Vercauteren. 2019.
\newblock Aleatoric uncertainty estimation with test-time augmentation for
  medical image segmentation with convolutional neural networks.
\newblock \emph{Neurocomputing}.

\bibitem[{Wang et~al.(2021)Wang, Yin, Yao, Zhang, Fu, Ding, Li, Huang, and
  Xue}]{Wang_2021_CVPR}
Wenxuan Wang, Bangjie Yin, Taiping Yao, Li~Zhang, Yanwei Fu, Shouhong Ding,
  Jilin Li, Feiyue Huang, and Xiangyang Xue. 2021.
\newblock Delving into data: Effectively substitute training for black-box
  attack.
\newblock In \emph{CVPR 2021}.

\bibitem[{Wang(2021)}]{wang2021dbkd}
Zi~Wang. 2021.
\newblock Zero-shot knowledge distillation from a decision-based black-box
  model.
\newblock In \emph{ICML 2021}.

\bibitem[{Warstadt et~al.(2019)Warstadt, Singh, and Bowman}]{cola}
Alex Warstadt, Amanpreet Singh, and Samuel Bowman. 2019.
\newblock Neural network acceptability judgments.
\newblock \emph{TACL 2019}.

\bibitem[{Wei and Zou(2019)}]{eda}
Jason Wei and Kai Zou. 2019.
\newblock {EDA}: Easy data augmentation techniques for boosting performance on
  text classification tasks.
\newblock In \emph{EMNLP 2019}.

\bibitem[{Wei et~al.(2022)Wei, Yu, Hu, Weng, Luo, and Jin}]{wei2022learningaug}
Xiangpeng Wei, Heng Yu, Yue Hu, Rongxiang Weng, Weihua Luo, and Rong Jin. 2022.
\newblock Learning to generalize to more: Continuous semantic augmentation for
  neural machine translation.
\newblock In \emph{ACL 2022}.

\bibitem[{Williams et~al.(2018)Williams, Nangia, and Bowman}]{mnli}
Adina Williams, Nikita Nangia, and Samuel Bowman. 2018.
\newblock A broad-coverage challenge corpus for sentence understanding through
  inference.
\newblock In \emph{NAACL 2018}.

\bibitem[{Wolf et~al.(2020)Wolf, Debut, Sanh, Chaumond, Delangue, Moi, Cistac,
  Rault, Louf, Funtowicz, Davison, Shleifer, von Platen, Ma, Jernite, Plu, Xu,
  Le~Scao, Gugger, Drame, Lhoest, and Rush}]{huggingface}
Thomas Wolf, Lysandre Debut, Victor Sanh, Julien Chaumond, Clement Delangue,
  Anthony Moi, Pierric Cistac, Tim Rault, Remi Louf, Morgan Funtowicz, Joe
  Davison, Sam Shleifer, Patrick von Platen, Clara Ma, Yacine Jernite, Julien
  Plu, Canwen Xu, Teven Le~Scao, Sylvain Gugger, Mariama Drame, Quentin Lhoest,
  and Alexander Rush. 2020.
\newblock Transformers: State-of-the-art natural language processing.
\newblock In \emph{EMNLP 2020: System Demonstrations}.

\bibitem[{Xu et~al.(2022)Xu, Berti-Equille, Cuesta-Infante, and
  Veeramachaneni}]{xu2022situ}
Lei Xu, Laure Berti-Equille, Alfredo Cuesta-Infante, and Kalyan Veeramachaneni.
  2022.
\newblock In situ augmentation for defending against adversarial attacks on
  text classifiers.
\newblock In \emph{KDD 2022 Workshop on Adversarial Learning Methods for
  Machine Learning and Data Mining}.

\bibitem[{Yang et~al.(2020)Yang, Malaviya, Fernandez, Swayamdipta, Le~Bras,
  Wang, Bhagavatula, Choi, and Downey}]{yang2020generativeaug}
Yiben Yang, Chaitanya Malaviya, Jared Fernandez, Swabha Swayamdipta, Ronan
  Le~Bras, Ji-Ping Wang, Chandra Bhagavatula, Yejin Choi, and Doug Downey.
  2020.
\newblock Generative data augmentation for commonsense reasoning.
\newblock In \emph{Findings of the EMNLP 2020}.

\bibitem[{Yu and Sun(2022)}]{Yu-2022-FE-DaST}
Mengran Yu and Shiliang Sun. 2022.
\newblock {FE-DaST}: Fast and effective data-free substitute training for
  black-box adversarial attacks.
\newblock \emph{Computers \& Security}.

\bibitem[{Zhang et~al.(2022)Zhang, Chen, Dong, Jia, and Lyu}]{qekd}
Jie Zhang, Chen Chen, Jiahua Dong, Ruoxi Jia, and Lingjuan Lyu. 2022.
\newblock {QEKD}: Query-efficient and data-free knowledge distillation from
  black-box models.

\bibitem[{Zhou et~al.(2020)Zhou, Wu, Liu, Liu, and Zhu}]{Zhou_2020_CVPR}
Mingyi Zhou, Jing Wu, Yipeng Liu, Shuaicheng Liu, and Ce~Zhu. 2020.
\newblock {DaST}: Data-free substitute training for adversarial attacks.
\newblock In \emph{CVPR 2020}.

\end{thebibliography}
